\documentclass[11pt]{article}
\usepackage{acl-hlt2011}
\usepackage{times}
\usepackage{latexsym}
\usepackage{amsmath}
\usepackage{multirow}
\usepackage{url}

\usepackage{graphicx}
\usepackage{tabularx}
\usepackage{siunitx}
\usepackage{microtype}
\usepackage{booktabs}
\usepackage{soul}

\title{The University of Texas at Dallas HLTRI's Participation in EPIC-QA: Searching for Entailed Questions Revealing Novel Answer Nuggets}

\author{Maxwell Weinzierl \\
  Human Language Technology \\ 
  Research Institute,  \\
  University of Texas at Dallas \\
  \texttt{maxwell.weinzierl@utdallas.edu} \\
  \And
  Sanda M. Harabagiu \\
  Human Language Technology \\
  Research Institute, \\
  University of Texas at Dallas \\
  \texttt{sanda@utdallas.edu} \\
}

\date{}

\begin{document}
\maketitle
\begin{abstract}
The  Epidemic Question Answering (EPIC-QA) track at the Text Analysis Conference (TAC) is an evaluation of methodologies for answering ad-hoc questions about the COVID-19 disease. 
This paper describes our participation in both tasks of EPIC-QA, targeting: (1) Expert QA and (2) Consumer QA. Our methods used a multi-phase neural Information Retrieval (IR) system based on combining BM25, BERT, and T5 as well as the idea of considering {\em entailment relations} between the original question and questions automatically generated from answer candidate sentences. Moreover, because entailment relations were also considered between all generated questions, we were able to re-rank the answer sentences based on the number of novel answer nuggets they contained, as indicated by the processing of a {\em question entailment graph}. Our system, called SEaRching for Entailed QUestions revealing NOVel nuggets of Answers (SER4EQUNOVA), produced promising results in both EPIC-QA tasks, excelling in the Expert QA task.  

\end{abstract}

\section{Introduction}
Answering questions related to the COVID-19 pandemic can inform both medical experts and the general public, referred to as “consumers” of biomedical and public health information. 
Interestingly, and not surprising, the questions asked by these two types of users of automated Question/Answering (Q/A) systems vary significantly in both terminology, level of detail, and purpose. 
Experts include clinicians, biomedical researchers, healthcare workers and well as public health experts, while consumers include average citizens, patients, and governmental policymakers. 
The information needs of these user groups can vary drastically, therefore audience-specific corpora and question-answering systems were constructed by the organizers of the Epidemic Question Answering (EPIC-QA) track at TAC 2020, to satisfy these needs. 
More specifically, the tasks of EPIC-QA were: \\
$\Box$ \underline{Task A: Expert QA}: In Task A, teams were provided with a set of questions asked by experts, being required to provide a ranked list of expert-level answers to each question, with the understanding that the answers should provide information that is useful to researchers, scientists, or clinicians.\\
$\Box$  \underline{Task B: Consumer QA}: In Task B, teams were provided with a set of questions asked by consumers, being required to provide a ranked list of consumer-friendly answers to each question, with the constrain that answers should be understandable by the general public.\\
In both EPIC-QA tasks, the answer format consists of a set of consecutive sentences extracted from a single context of a single document. Contexts and sentence IDs were provided to the EPIC-QA participants. These contexts and sentences originated in task-specific corpora. Each task had a different corpus. 

The Task A of EPIC-QA used the collection of biomedical articles released for the COVID-19 Open Research Dataset Challenge (CORD-19), which includes a subset of articles in PubMed Central (PMC) as well as pre-prints from bioRxiv. Contexts in this collection correspond to automatically identified paragraphs in the articles' abstracts, or main texts.

The Task B of EPIC-QA used a subset of the articles used by the Consumer Health Information Question Answering (CHIQA) service of the U.S. National Library of Medicine (NLM). This collection includes authoritative articles from: the Centers for Disease Control and Prevention (CDC); the Genetic and Rare Disease Information Center (GARD); the Genetics Home Reference (GHR); Medline Plus; the National Institute of Allergy and Infectious Diseases (NIAID); the World Health Organization (WHO); Contexts in this collection correspond to paragraphs or sections as indicated by the HTML markup of the document. In addition,
265 reddit threads from /r/askscience tagged with COVID-19, Medicine, Biology, or the Human Body, and filtered for COVID-19 content were used in this task. Furthermore, a subset of the CommonCrawl News crawl from January 1st to April 30th, 2020, as used in the TREC Health Misinformation Track were considered after filtering them by domain using SALSA, PageRank, and HITS and further filtered them for COVID-19 content.

Given that both EPIC-QA tasks are essentially Question/Answering (Q/A) tasks, we believed from the beginning that we should develop a single Q/A system. Moreover, we believed that the system should be capable to both rank the candidate answers and to be able to identify multiple, distinct atomic facts (called {\em answer nuggets}) that answer a question. While ranking can be achieved through a variety of relevance methods, the discovery of the answer nuggets should result from some form of inference between the question and the snippets of texts from the answer. Such an inference is achievable from the {\em textual entailment} relation. But we were convinced that the ideal inference should take place between the original question and the questions that we could generate automatically from candidate answers. Therefore, at the core of our methodology stays a search problem, which finds the distinct answer nuggets by searching for those questions automatically generated from candidate answers that are entailed by the original question. This informs a re-ranking of the candidate answers, to prioritize answers that generated most of the entailed questions. This enabled us to design the system for SEaRching for Entailed QUestions revealing NOVel nuggets of Answers (SER4EQUNOVA), which is described in this paper. 

\begin{figure}[ht]
    \centering
    \includegraphics[width=0.48\textwidth]{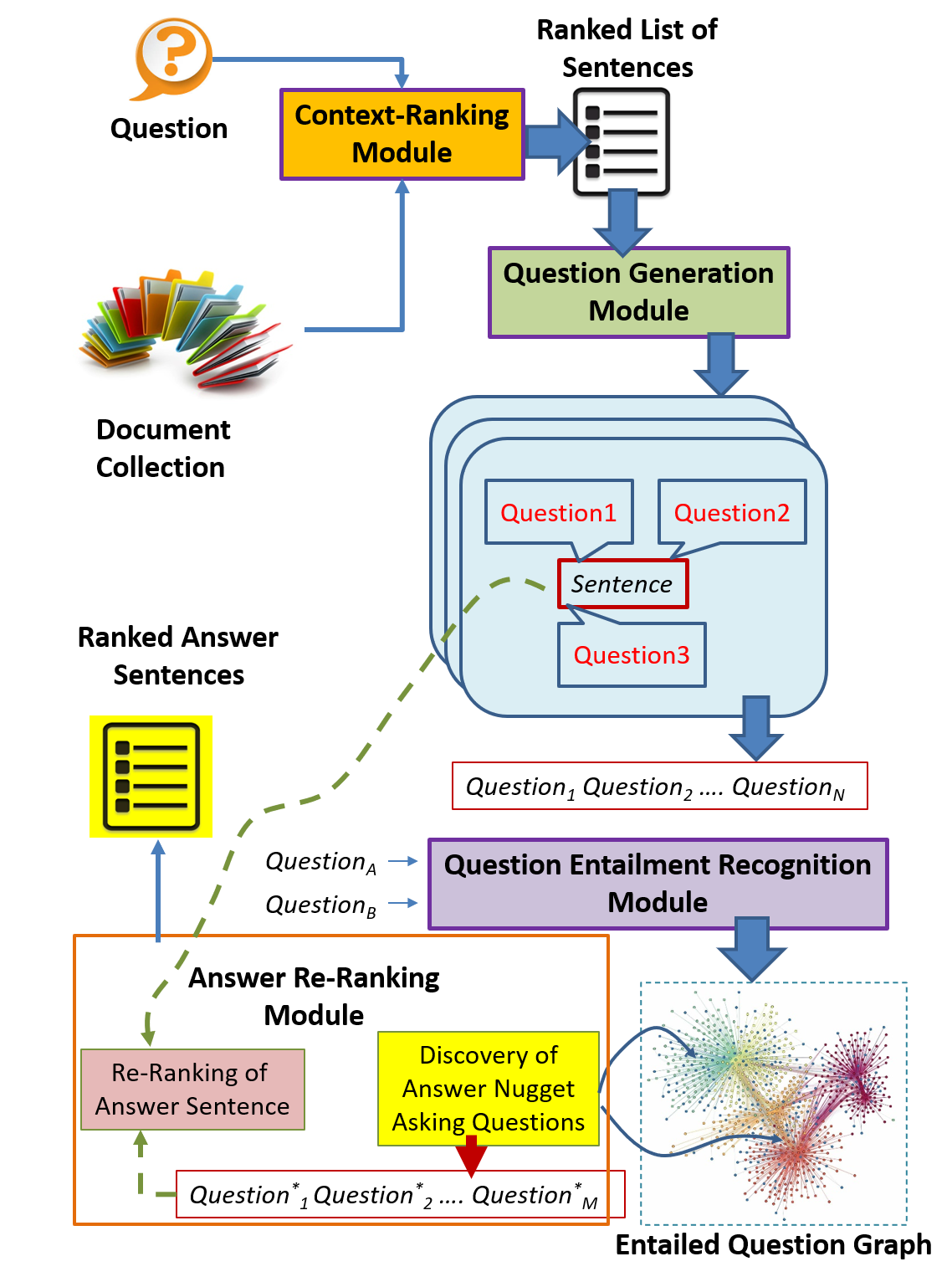}
    \caption{The Architecture of the System for SEaRching for Entailed QUestions revealing NOVel nuggets of Answers (SER4EQUNOVA).}
    \label{fig:architecture}
\end{figure}

\section{The Architecture}

\begin{figure*}[ht]
    \centering
    \includegraphics[width=0.8\textwidth]{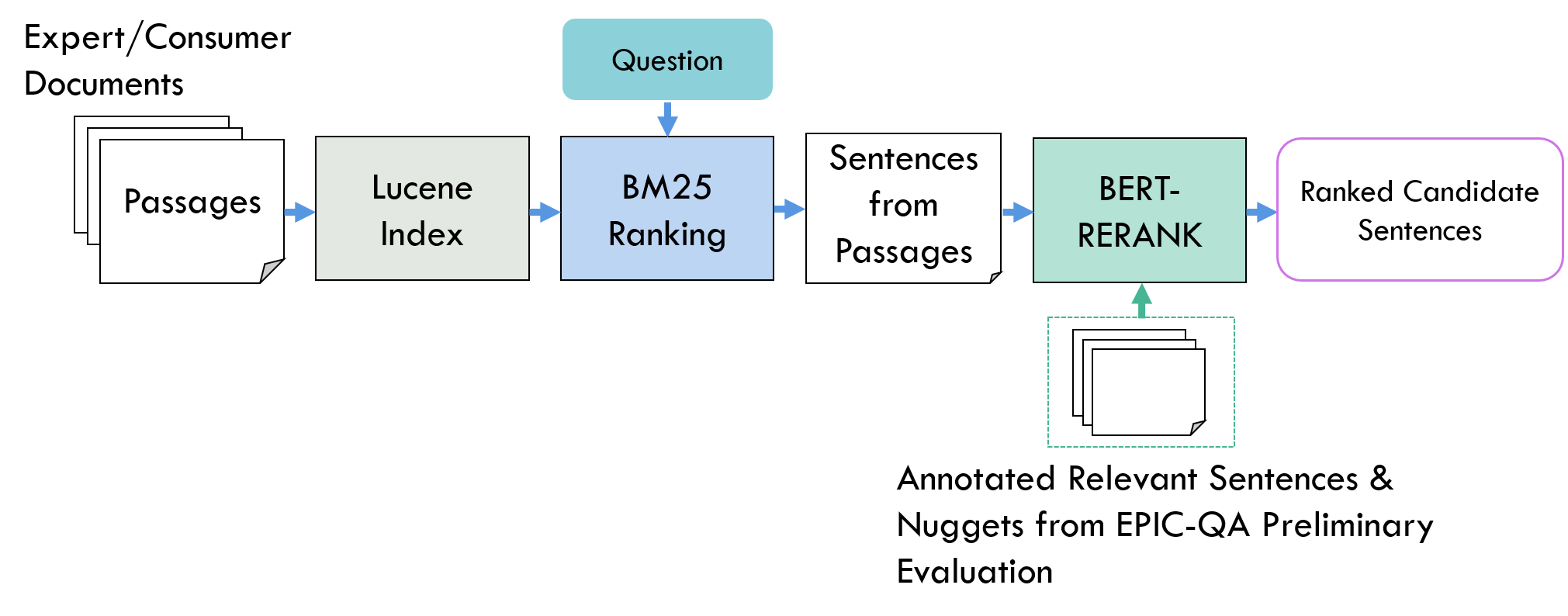}
    \caption{Context-Ranking Module of SER4EQUNOVA.}
    \label{fig:answer_retrieval_module}
\end{figure*}

Figure~\ref{fig:architecture} illustrates the architecture of the SER4EQUNOVA system. Given a question and the document collection used in one of the EPIC-QA tasks, the Context-Ranking Module retrieves ranked sentences from the document collection that are candidates for finding the answers expected by the question. From each such sentence, questions are generated by the Question Generation Module. Pairs of questions are passed through the Question Entailment Recognition module, to discover which question is textually entailed by another question. These pairs of questions consist either of questions that were generated from the candidate sentences, or the original question that is pairs with any of the generated questions. The discovered entailment relations enable the generation of the
Entailment Question Graph (EQG), where nodes are questions and vertices are informed by the discovered entailment relations. 

The Answer Re-Ranking Module, illustrated in Figure 2 has two functions. First, it discovered in the EGG those questions that ask about answer nuggets of the original question. This is made possible by discovering in the EQG the largest connected components. The questions that (1) are entailed by the original question and (2) have the largest connectivity in each connected component indicate that they are not only pinpointing to an answer of the original question, but that answer is a novel answer nugget.
Based on this assumption, the answer sentences are re-ranked, to favor those sentences that 
contain the largest number of answer nuggets. The re-ranked list of answer sentences is the output of the SER4EQUNOVA system.

\subsection{The Context-Ranking Module}

The first module of SER4EQUNOVA is the Context-Ranking Module, which is tasked with searching for relevant candidate answers for a given question from the Expert or Consumer collection. 
The Context-Ranking Module is illustrated in Figure~\ref{fig:answer_retrieval_module}. 
The corpus (Expert or Consumer) is indexed with Lucene and searched for relevant passages (called "contexts" in the shared-task) with BM25\cite{bm25}. 
Next, we considered all sentences within each context and produced a re-ranking score against each question using our BERT-RERANK system. 
We select the top-1000 sentences for each question from this ranking to pass on to the question generation module.

BERT-RERANK was trained similarly to prior work \cite{bert-rerank}: We consider sentences which contain a nugget for a given question as a "relevant" sentence, and we randomly sample "not\_relevant" sentences from within the top-1000 sentences from BM25. 
Training is performed on the "preliminary" collection provided by the track annotators. 
We trained separate models for both tasks A and B. 
We initialized the BERT weights to a re-ranking model which was trained on MSMARCO \cite{msmarco} using BioBERT \cite{BioBERT}, a biomedical domain-specific BERT language model. 
We additionally weighed the loss by multiplying the loss of "relevant" sentences by the number of nuggets within that sentence. 
This further prioritizes re-ranking sentences with more nuggets.

\subsection{The Question Generation Module}
\begin{figure*}[ht]
    \centering
    \includegraphics[width=0.8\textwidth]{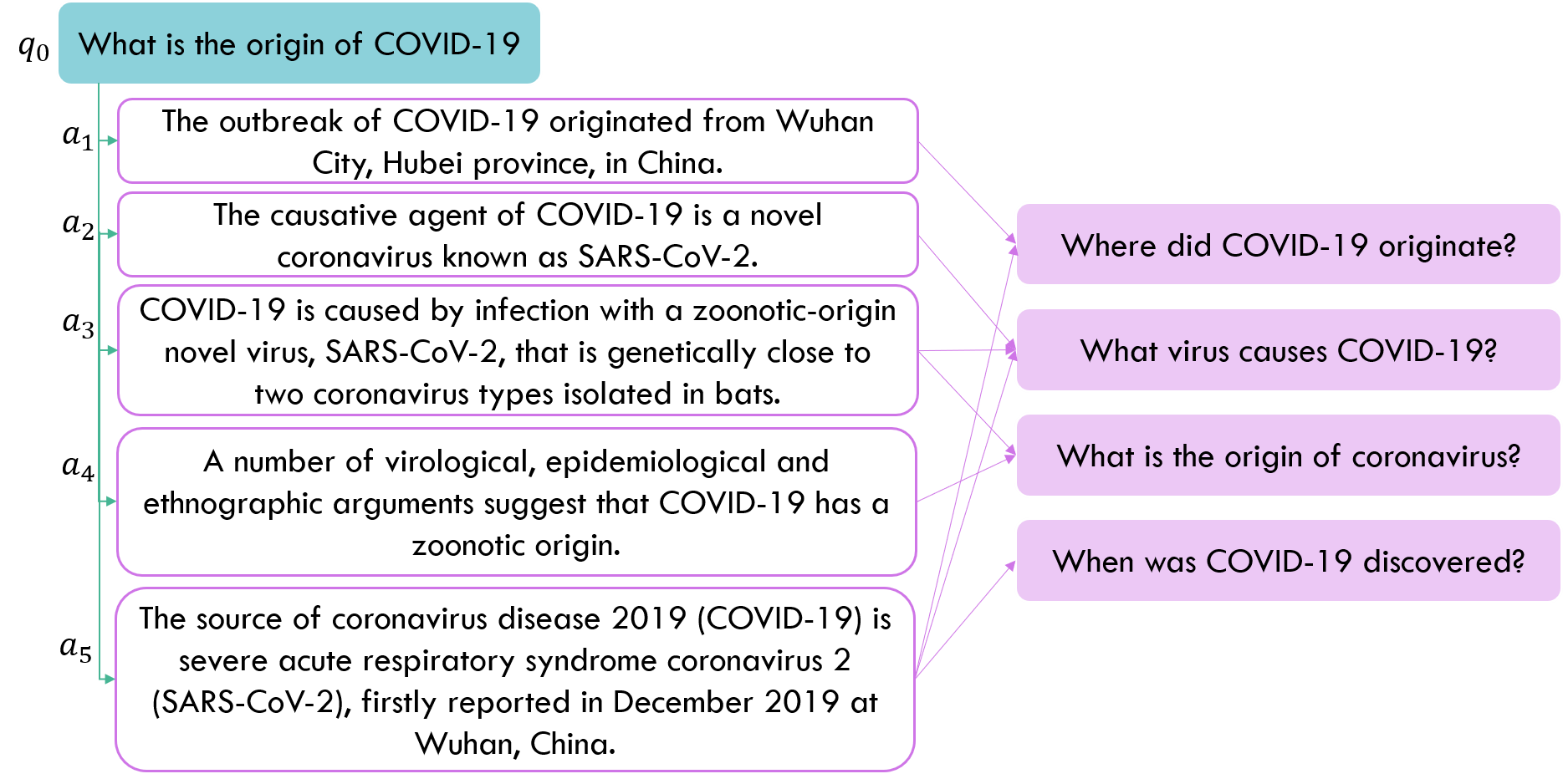}
    \caption{Example of Question used in EPIC-QA Task 1, its Retrieved Answer Sentences and  Generated New Questions.}
    \label{fig:example}
\end{figure*}
The Question Generation Module produces synthetic, generated questions\cite{doc-exp-query-pred} for every candidate answer provided by the Context-Ranking Module. Questions are generated with the docTTTTTquery model \cite{doct5query} trained on MSMARCO \cite{msmarco}, but in the future this system could be fine-tuned for biomedical question generation. For each candidate answer we generate $k$ questions.
Question generation can be performed entirely offline on the corpus, but it can also run ad-hoc as it is an embarrassingly parallel process with respect to each answer.
An example from the Expert QA collection is provided in Figure~\ref{fig:example}. 
For question $q_0$: "What is the origin of COVID-19" we illustrate a subset of the candidate answers produced by the Context-Ranking Module and a subset of generated questions for each answer. Each answer is connected to one or more questions which were generated from that answer. For example, the answer "The outbreak of COVID-19 originated from Wuhan City, Hubei province, in China." produced the question "Where did COVID-19 originate?". The method by which these generated questions are connected across answers is described next. 

\subsection{The Question Entailment Recognition Module}
\begin{figure}[ht]
    \centering
    \includegraphics[width=0.38\textwidth]{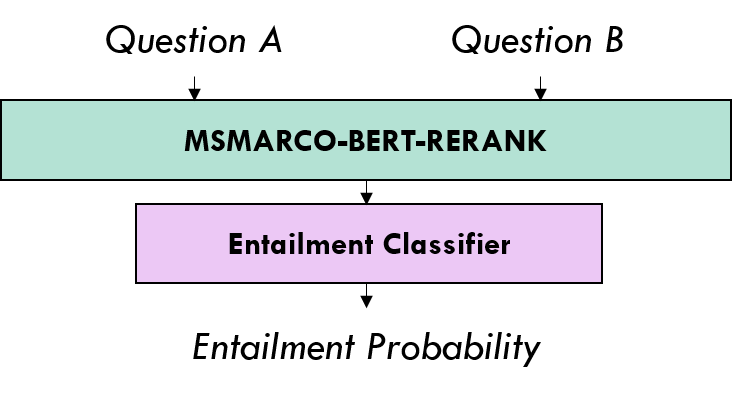}
    \caption{The Usage of QBERT-RQE in the Question Entailment Recognition Module.}
    \label{fig:rqe_architecture}
\end{figure}

The Entailed Question Graph requires a Recognizing Question Entailment (RQE)\cite{rqe-qa} system to identify question entailment relations between pairs of questions. 
We experimented with two RQE systems: BERT-RQE and QBERT-RQE. BERT-RQE, illustrated in Figure~\ref{fig:rqe_architecture}, utilizes the pre-trained BERT language model\cite{bert} with a single linear layer which predicts "entailed" or "not\_entailed" relations between question pairs. The second RQE system, QBERT-RQE utilizes the same architecture, but initializes its weights to those of MSMARCO-BERT-RERANK trained on question-answer pairs from MSMCARCO \cite{msmarco} to predict "relevant" or "not\_relevant"\cite{bert-rerank}, with the intuition being that question-answer relevancy is similar to question-question entailment. 

Both models were fine-tuned on the Quora question duplication dataset\footnote{https://www.kaggle.com/c/quora-question-pairs} and outperform other published models, as is reported in the Results section.
The word-piece tokens \cite{bert} for both question A and question B are provided together to MSMARCO-BERT-RERANK along with a sequence start token "[CLS]" and sequence separator tokens "[SEP]". This BERT-based model produces a single "contextualized" embedding for the "[CLS]" token, which is provided to a single entailment classification layer to produce the final entailment probability between question A and question B.

\subsection{The Answer Re-Ranking Module}
The Answer Re-Ranking Module is tasked with re-ranking the answers provided by the Context-Ranking Module based on the novelty of the nuggets within each answer. The questions generated by the Question Generation Module are utilized to construct an Entailed Question Graph, which is utilized to discover answer nugget asking questions. The presence of these answer nugget asking questions in answers provides the necessary nugget information for answer sentence re-ranking. The Answer Re-Ranking Module outputs a ranked list of answer sentences which rank higher answers which contain more novel nuggets relative to other answers. 

\subsubsection{Generating the Entailed Question Graph}

\begin{figure}[ht]
    \centering
    \includegraphics[width=0.48\textwidth]{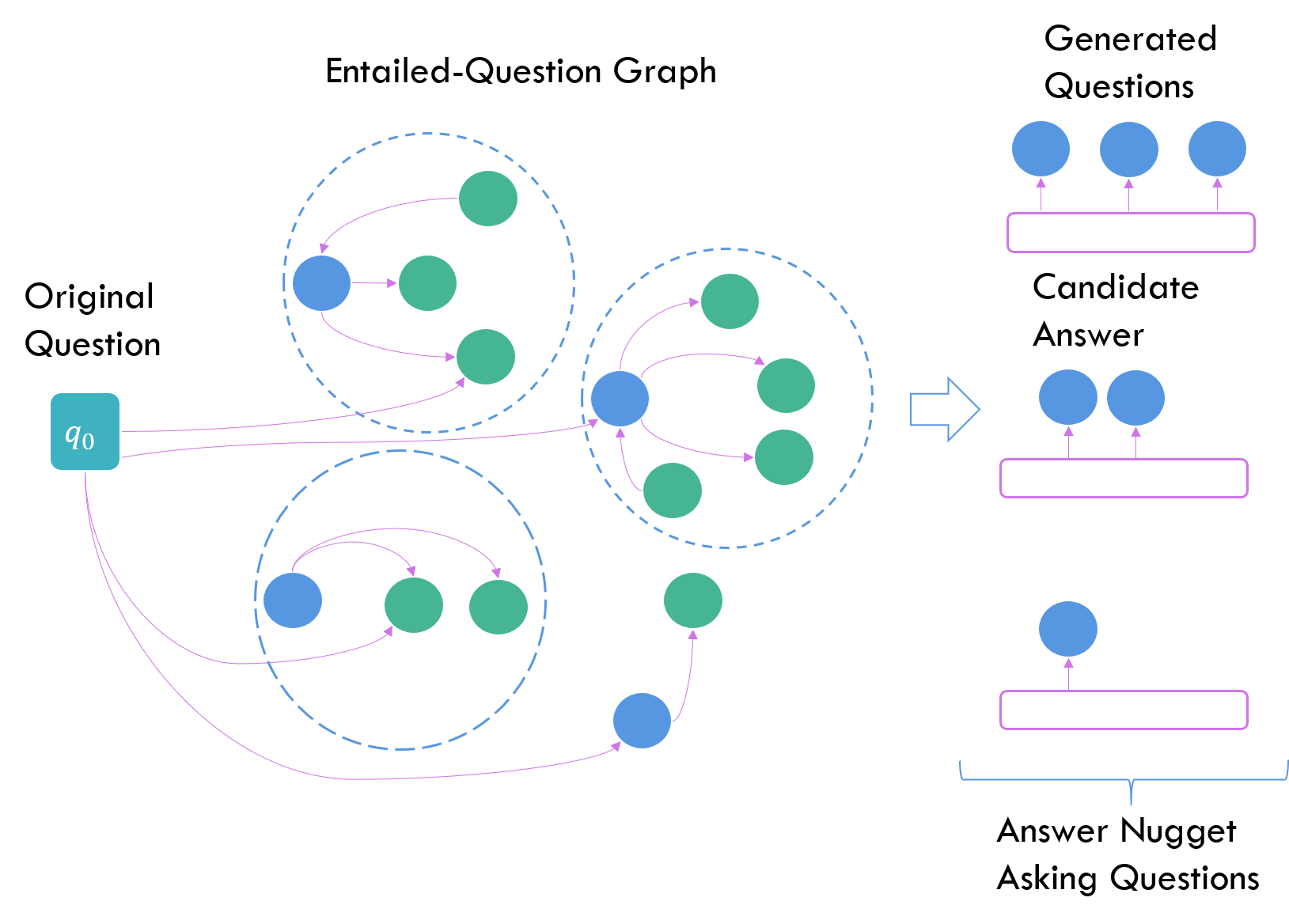}
    \caption{Entailment Question Graph}
    \label{fig:entailed_question_graph}
\end{figure}

The Entailed Question Graph is illustrated in Figure~\ref{fig:entailed_question_graph}. 
The generation of the Entailed Question Graph (EQG) relies on the prediction of the  entailment probability between pairs of questions. The nodes in the EQG are all the questions generated by the Question Generation Module which also share some entailment relation. It is to be noted that we consider the {\em original question} $q_0$
that is evaluated either in Task A or B as external to the EQG. Nevertheless, we also account for the entailment relations that are identified by the Question Entailment Recognition module between $q_0$ and any of the question nodes from the EQG, as illustrated in Figure~\ref{fig:entailed_question_graph}.

The EQG is used for discovering those automatically generated questions that ask for the answer nuggets of $q_0$, the original question. To discover those questions, we first find all the connected components of the EQG, i.e. set of connected questions that are spanned by entailment relations, as illustrated in Figure~\ref{fig:entailed_question_graph}. From each connected component of the EQG, we select those questions that (1) are connected through the largest number of entailment relations to other questions from the same component; and (2) are also entailed by $q_0$, In Figure~\ref{fig:entailed_question_graph}, we distinguish these questions by representing them as blue nodes in the EQG. We further filter out from these selected questions those that have an entailment probability from  $q_0$ lower than a pre-set entailment threshold. We believe that this would allow us to consider only automatically-generated questions that have a good likelihood to ask for various answer nuggets of $q_0$. We denote these questions as Answer Nugget Asking Questions.

\begin{table*}[h]
\centering
\small
\begin{tabular}{lccc}
    \toprule
    {\sc Expert}               & {\bf NDNS-Relaxed}  & {\bf NDNS-Partial}   & {\bf NDNS-Exact}  \\
    \midrule
    Best  & \bf{0.371} & \bf{0.370} & \bf{0.421}\\
    Median & 0.339 & 0.338 & 0.380\\
    \hline  
    CRM (RUN\_2) & 0.364 & 0.363 & 0.413\\
    SER4EQUNOVA (RUN\_3) & \bf{0.371} & \bf{0.370} & \bf{0.421}\\
    \midrule
    {\sc Consumer}             & {\bf NDNS-Relaxed} & {\bf NDNS-Partial}    & {\bf NDNS-Exact}  \\
    \midrule
    Best  & \bf{0.368} & {\bf 0.366} & \bf{0.414} \\
    Median  & 0.284 & 0.286 & 0.314 \\
    \hline  
    CRM (RUN\_2) & 0.313 & 0.312 & 0.353 \\
    SER4EQUNOVA (RUN\_3) & 0.317 & 0.316 & 0.363 \\
    \bottomrule
\end{tabular}
\caption{Results on the Task A and B of the 2020 TAC EPIC-QA, measured using  NDNS.}
\label{tb:results}
\end{table*}

\subsubsection{Re-Ranking of Answer Sentences}
The Answer Re-Ranking Module takes into account the answer nugget asking questions and re-ranks the answer sentences retrieved by the Context Ranking Module. 
The idea is not only to count the number of answer nugget asking questions that were generated from each answer sentence but also to account for the {\em novelty} of the answer nuggets. That entails keeping track of each answer nugget that is already known and distinguish the novel answer nuggets. It is to be noted that when questions are automatically generated in the Question Generation Module, for each question we also know the text snippet for the answer sentence that responds to the newly generated question. In the case of the answer nugget asking questions, we keep track of each corresponding answer snippet that answers them, considering possible answer nuggets for $q_0$. 

The idea is that each answer sentence from which answer nugget asking questions were generated may contain either "known" answer nuggets, which have been observed in other answer sentences, or "novel" answer nuggets, yet not seen anywhere else. 
Hence, it is ideal to prefer answer sentences which contain nuggets that have little or no overlap with previous seen answer nuggets. Taking this into account,  we modified the ranking produced by the Context-Ranking Module by  making use of the same NDNS algorithm utilized by the EPIC-QA organizers to compute an optimal ranking of answers based on nuggets within each answer. 
However, we substituted the answer nuggets with the answer nugget asking questions, but the algorithm operates exactly the same otherwise. 

For example, the re-ranking of the example in Figure~\ref{fig:example} would be the following:
The question $q_0$ "What is the origin of COVID-19" would be provided to our Context-Ranking Module, which would result in five candidate answers ranked by relevancy: $a_1, a_2, a_3, a_4,$ and $a_5$. 
The Question Generation Module would generate $k$ questions for each answer, and the Question Entailment Recognition Module would produce four answer nugget asking questions. 
Re-Ranking of Answer Sentences utilizes these answer nugget asking questions to infer the presence of nuggets, which is used to re-rank the answers. 
The NDNS algorithm would select $a_5$ first, since it contains 3 novel nuggets. 
The next answer would be $a_3$ since it contains the only remaining unseen nugget and $a_3$ ranks above $a_4$. 
Finally, as all nuggets have been seen we keep the remaining relative ordering of $a_1, a_2, $ and $a_4$. 

\section{Results}

The performance of the SER4EQUNOVA system on the Task A and B was evaluated by the EPIC-QA  organizers. Our team submitted two primary runs, with the third run containing a test run not reported in this paper. 
Our first run "RUN\_2" utilized only the Context-Ranking Module (CRM) of SER4EQUNOVA, while the second run "RUN\_3" utilized the entire system. 

The evaluation metric was based on a modified form of Normalized Discounted Cumulative Gain (NDCG) called Normalized Discount Novelty Score (NDNS). NDNS utilizes the following novelty score $NS_i$ for an answer $i$:

\begin{equation}
    NS_i =  \frac{\#_i \times (\#_i + 1)}{\#_i + SF_i}
\end{equation}

where $\#_i$ is the number of novel nuggets in an answer $i$ and $SF_i$ is a sentence factor which penalizes the novelty score. A nugget is considered novel if it has not been present in an answer retrieved earlier in the ranked list. The sentence factor $SF_i$ is different for each of the three NDNS metrics considered:
\begin{itemize}
    \item NDNS-Relaxed: Answers should only contain sentences with novel nuggets. $SF_i = \#na_i + \#sn_i + min(\#nn_i, 1)$
    \item NDNS-Partial: Answers should not contain sentences with no nuggets. $SF_i = \#na_i + min(\#nn_i, 1)$
    \item NDNS-Exact: Answers should be short. $SF_i = \#na_i + \#sn_i + \#nn_i$.
\end{itemize}
where $\#na_i$ is the number of sentences with no nuggets, $\#sn_i$ is the number of sentences with seen nuggets, and $\#nn_i$ is the number of sentences with novel nuggets.
The novelty score $NS_i$ is used as the cumulative gain for adding a new answer to a list of previously added answers, and the Normalized Discount Cumulative Gain algorithm uses this answer score to produce a score for an entire ranked list of answers. 

Results from the final evaluation of EPIC-QA are available in Table~\ref{tb:results}.

\begin{table}[ht]
\centering
\small
\begin{tabular}{lc}
    \toprule
    System & {\bf Accuracy (\%)}   \\
    \midrule
    Logistic Regression + Features  & 67.79 \\
    NN  & 81.34 \\
    NN + GloVe embeddings & 83.62 \\
    \hline  
    BERT-RQE & 88.94 \\
    QBERT-RQE & \bf{89.55} \\
    \bottomrule
\end{tabular}
\caption{Results of Question Entailment Recognition on the Quora Question Duplication dataset.}
\label{tb:quora-results}
\end{table}

Because the SER4EQUNOVA system makes use of the Recognizing Question Entailment (RQE) module, we also performed an comparative evaluation of the two implementations we have considered for this module and well as several baselines. To evaluate the performance if the question entailment, we used the the test collection of the Quora Question Duplication dataset. The results of both QBERT-RQE, a state-of-the-art Recognizing Question Entailment (RQE) method, and BERT-RQE on this test collection are provided in Table~\ref{tb:quora-results}. We can also see  that utilizing the question-answer re-ranking model MSMARCO-BERT-RERANK improves question-question entailment performance over simply using the pre-trained BERT.

\section{Conclusion}
In this work we present the SEaRching for Entailed QUestions revealing NOVel nuggets of Answers (SER4EQUNOVA) system for answering ad-hoc questions about the COVID-19 disease for both Expert and Consumer audiences. SER4EQUNOVA uses a multi-phase neural Information Retrieval (IR) system which identifies and re-ranks answers based on the presence of novel nuggets by considering the {\em entailment relations} between the original question and questions automatically generated from answer candidate sentences. The results indicate that this {\em question entailment graph} approach is beneficial for answering questions when the novelty of the factual nuggets present in answers presented to the user is central to the task.

\bibliography{acl-hlt2011}

\begin{thebibliography}{}

\bibitem[\protect\citename{Ben~Abacha and Demner-Fushman}2019]{rqe-qa}
Asma Ben~Abacha and Dina Demner-Fushman.
\newblock 2019.
\newblock A question-entailment approach to question answering.
\newblock {\em BMC Bioinformatics}, 20:511, 10.

\bibitem[\protect\citename{Devlin \bgroup et al.\egroup }2019]{bert}
Jacob Devlin, Ming-Wei Chang, Kenton Lee, and Kristina Toutanova.
\newblock 2019.
\newblock {BERT}: Pre-training of deep bidirectional transformers for language
  understanding.
\newblock In {\em Proceedings of the 2019 Conference of the North {A}merican
  Chapter of the Association for Computational Linguistics: Human Language
  Technologies, Volume 1 (Long and Short Papers)}, pages 4171--4186.
  Association for Computational Linguistics.

\bibitem[\protect\citename{Lee \bgroup et al.\egroup }2019]{BioBERT}
Jinhyuk Lee, Wonjin Yoon, Sungdong Kim, Donghyeon Kim, Sunkyu Kim, Chan~Ho So,
  and Jaewoo Kang.
\newblock 2019.
\newblock Biobert: a pre-trained biomedical language representation model for
  biomedical text mining.
\newblock {\em Bioinformatics}.

\bibitem[\protect\citename{Nguyen \bgroup et al.\egroup }2016]{msmarco}
Tri Nguyen, Mir Rosenberg, Xia Song, Jianfeng Gao, Saurabh Tiwary, Rangan
  Majumder, and Li~Deng.
\newblock 2016.
\newblock {MS} {MARCO:} {A} human generated machine reading comprehension
  dataset.
\newblock {\em CoRR}, abs/1611.09268.

\bibitem[\protect\citename{Nogueira and Cho}2020]{bert-rerank}
Rodrigo Nogueira and Kyunghyun Cho.
\newblock 2020.
\newblock Passage re-ranking with bert.

\bibitem[\protect\citename{Nogueira \bgroup et al.\egroup
  }2019]{doc-exp-query-pred}
Rodrigo Nogueira, Wei Yang, Jimmy Lin, and Kyunghyun Cho.
\newblock 2019.
\newblock Document expansion by query prediction.
\newblock {\em CoRR}, abs/1904.08375.

\bibitem[\protect\citename{Nogueira}2019]{doct5query}
Rodrigo Nogueira.
\newblock 2019.
\newblock From doc2query to doctttttquery.

\bibitem[\protect\citename{Robertson \bgroup et al.\egroup }1994]{bm25}
Stephen Robertson, Steve Walker, Susan Jones, Micheline Hancock-Beaulieu, and
  Mike Gatford.
\newblock 1994.
\newblock Okapi at trec-3.
\newblock pages 0--, 01.

\end{thebibliography}
\bibliographystyle{acl}

\end{document}